\documentclass[conference]{IEEEtran}
\IEEEoverridecommandlockouts
\usepackage{cite}
\usepackage{amsmath,amssymb,amsfonts}
\usepackage{algorithmic}
\usepackage{graphicx}
\usepackage{textcomp}
\usepackage{xcolor}
\usepackage{caption} 
\usepackage{graphicx} 
\usepackage{array}
\usepackage{listings}
\usepackage{tikz}
\usepackage{booktabs}
\usepackage{tabularx}
\usetikzlibrary{positioning}
\usetikzlibrary{positioning}
\usepackage[utf8]{inputenc}
\def\BibTeX{{\rm B\kern-.05em{\sc i\kern-.025em b}\kern-.08em
    T\kern-.1667em\lower.7ex\hbox{E}\kern-.125emX}}
\begin{document}

\title{Rule-Based Moral Principles for Explaining Uncertainty in Natural Language Generation\\

}

\author{\IEEEauthorblockN{1\textsuperscript{st} Zahra Atf}
\IEEEauthorblockA{\textit{Faculty of Business and Information Technology} \\
\textit{Ontario Tech University}\\
Oshawa, Canada \\
0000-0003-0642-4341}
\and
\IEEEauthorblockN{2\textsuperscript{nd} Peter R. Lewis}
\IEEEauthorblockA{\textit{Faculty of Business and Information Technology} \\
\textit{Ontario Tech University}\\
Oshawa, Canada \\
0000-0003-4271-8611}
}



\maketitle

\begin{abstract}
Rule-Based Moral Principles for Explaining Uncertainty in Natural Language Generation
As large language models (LLMs) are increasingly used in high-stakes applications, the challenge of explaining  uncertainty in natural language generation has become both a technical and moral imperative. Traditional approaches rely on probabilistic methods that are often opaque, difficult to interpret, and misaligned with human expectations of transparency and accountability. In response to these limitations,  this paper introduces a novel framework based on rule-based moral principles—simple, human-inspired ethical guidelines—for responding to uncertainty in LLM-generated text. Drawing on insights from experimental moral psychology and virtue ethics, we define a set of symbolic behavioral rules such as precaution, deference, and responsibility to guide system responses under conditions of epistemic or aleatoric uncertainty. These rules are implemented declaratively and are designed to generate adaptive, context-sensitive explanations even in the absence of precise confidence metrics.
The moral principles are encoded as symbolic rules within a lightweight Prolog-based engine, where each uncertainty tag (low, medium, high) activates an ethically aligned system action along with an automatically generated, plain-language rationale. We evaluate the framework through scenario-based simulations that benchmark rule coverage, assess fairness implications, and analyze trust calibration. An interpretive explanation module is integrated to reveal both the assigned uncertainty level and its underlying justification in a transparent and accessible way.
We illustrate the framework through hypothetical yet plausible use cases in clinical and legal domains, demonstrating how rule-based moral reasoning can enhance user trust, promote fairness, and improve the interpretability of AI-generated language. By offering a lightweight, philosophically grounded alternative to probabilistic uncertainty modeling, our approach paves the way for more ethical, human-aligned, and socially responsible natural language generation.
\end{abstract}


\section{Introduction}
Large Language Models (LLMs) have significantly advanced natural language processing by enabling the generation of coherent and contextually appropriate text across a wide range of domains \cite{Patsakis2024}. Their adoption in high-stakes applications—such as healthcare diagnostics \cite{Ghaffarzadeh2024}, legal reasoning \cite{Nay2024,AthulSaxena2024}, and financial forecasting \cite{Ma2024,Gregory2024}—has intensified the demand for robust uncertainty-management strategies to ensure system reliability and support informed decision-making.Explaining uncertainty entails quantifying, interpreting, and transparently conveying the confidence level of model predictions, thereby enabling more effective risk assessment and fostering user trust.\cite{gal2016dropout}.

Despite their impressive capabilities, current-generation LLMs such as GPT-3 and GPT-4 offer only limited access to internal signals, including token-level log-probabilities, temperature settings, and sampling distributions \cite{Amsterdam2024,openai2020gpt3}. These proxies are often insufficient for precise and interpretable uncertainty estimation. Decades of cognitive-psychology research further show that \emph{even when reliable probabilities are available, humans systematically misinterpret them}. Classic studies on heuristics and biases reveal pervasive anchoring, base-rate neglect, and miscalibration in probabilistic reasoning \cite{Zhang2023,Lieder2020}. Gigerenzer argues that people reason far more accurately with qualitative cues or natural frequencies than with abstract percentages \cite{Hertwig2021}. More recently, Sundh reviews behavioural responses to low-probability, high-impact events and reaches the same conclusion, a point echoed in contemporary popular analyses \cite{Sundh2024,BigThink2024}. These findings motivate our shift from numeric confidence scores to discrete uncertainty categories accompanied by plain-language moral rationales. Prior efforts to address estimation limits have included ensemble methods, Bayesian neural networks, and Monte Carlo dropout techniques as external uncertainty estimators \cite{Lakshminarayanan2017,Gal2016}. Simultaneously, explainability methods have aimed to expose model logic \cite{Ribeiro2016}, yet they often remain opaque, computationally expensive, or inaccessible to end-users.

Beyond technical challenges, the lack of transparent uncertainty communication raises important ethical concerns. Inadequate handling of uncertainty can exacerbate algorithmic bias, leading to unfair or discriminatory outcomes that disproportionately impact marginalized populations \cite{Bolukbasi2016} \cite{Atf2025}. These concerns are particularly acute in high-risk domains such as medicine, where even slight variations in prompts or model settings can result in dramatically different—and potentially unsafe—recommendations \cite{Savage2024,Lin2024}. As such, LLMs must not only generate plausible outputs but also signal when their confidence is low, especially in situations involving user safety and accountability \cite{Sheng2024,Feng2021}. Selective-classification and reject-option paradigms have been proposed to address this challenge formally \cite{El-Yaniv2010,liang2023}, but these solutions still rely heavily on opaque probabilistic models.

To address these limitations, we propose a rule-based framework that incorporates human-inspired moral principles into the uncertainty-response logic of LLMs. Instead of relying on numerical probability scores, our approach classifies model outputs into three qualitative uncertainty levels—\texttt{low}, \texttt{medium}, and \texttt{high}—and maps each level to a symbolic ethical rule. These rules operationalize three foundational principles: precaution (acting conservatively under high uncertainty), deference (referring ambiguous cases to human experts), and responsibility (acknowledging uncertainty and the limitations of automated systems). Each moral rule is accompanied by a natural-language explanation designed to enhance interpretability and calibrate user trust.

The primary aims of this research are threefold: (i) to formulate a minimal set of symbolic moral rules that guide system behavior under different uncertainty conditions; (ii) to implement these rules in a lightweight Prolog-based reasoning engine with an integrated explanation module; and (iii) to evaluate the framework through scenario-based simulations in clinical and legal domains, focusing on rule coverage, fairness implications, and trust calibration.

Accordingly, we frame our study around the following research questions:
\begin{itemize}
\item \textbf{RQ1:} Can a small set of moral rules serve as an effective alternative to opaque probabilistic estimates for signaling uncertainty?
\item \textbf{RQ2:} Does pairing each rule with a clear and concise natural-language explanation improve user trust and calibration?
\end{itemize}
A central component of our approach is a nuanced understanding of uncertainty itself. In AI and machine learning, uncertainty is commonly categorized into two types: \textit{epistemic uncertainty}, which arises from incomplete knowledge or model limitations, and \textit{aleatoric uncertainty}, which reflects intrinsic randomness in the data \cite{kendall2017}. We also consider a third category—\textit{ontological uncertainty}—which stems from the fundamental unpredictability of complex real-world phenomena and may remain unknowable in principle \cite{Dlugatch2024}. These categories frequently overlap in practice, particularly in clinical contexts where biological variability (aleatoric) coexists with data sparsity or ambiguity (epistemic) \cite{CASTANEDA202217,Hosseini2023}.\\
Taken together, the foregoing discussion establishes both the practical necessity and the cognitive–ethical motivation for a symbolic, virtue-aligned response layer.  The rest of the paper is organised as follows: Section 2 formalises our rule set and the underlying moral rationale; Section 3 details the Prolog-based implementation and its integration with a modern LLM; Section 4 presents scenario-based evaluations that quantify technical validity, fairness, and user-centred readability; and Section 5 concludes by reflecting on philosophical implications and outlining directions for entropy-aware tagging, cross-cultural virtue tuning, and formal verification of rule compliance.

\section{Related Work}
Cole et al.\ show that large language models can abstain on ambiguous questions by exploiting repetition patterns in sampled outputs rather than relying on soft-max probabilities \cite{Cole2023}.  Kamath, Jia, and Liang extend this idea under domain shift and train a calibrator that answers only 56 \% of questions while preserving 80 \% accuracy \cite{Kamath2020}.  These results motivate the precaution rule in our framework; however, our system turns abstention into a virtue-based obligation and accompanies every defer-or-warn action with a plain-language moral rationale, whereas the prior work uses learned numerical thresholds without ethical justification.

Desai and Durrett reduce over-confidence through temperature scaling and evaluate calibration error across domains \cite{Desai2020}.  ConU applies conformal prediction to black-box LLMs and derives prediction sets with correctness-coverage guarantees \cite{Wang2025COPU}, while COPU augments candidate outputs with the ground truth and measures non-conformity by logit scores to achieve reliable error control across a wide spectrum of user-specified rates \cite{Wang2025COPU}.  Our work differs by replacing distribution-free statistical guarantees with virtue-aligned behavioural commitments that communicate epistemic uncertainty in natural language rather than as set sizes or coverage numbers.

Pan et al.\ introduce Logic-LM, which converts natural-language problems into symbolic formulae before invoking a deterministic solver, yielding a 39 \% improvement on reasoning benchmarks \cite{Pan2023}.  d’Ascoli et al.\ show that Transformers can rediscover closed-form recurrence relations, bridging neural sequence models and symbolic regression \cite{dAscoli2022}.  Surveys by Bhuyan et al.\ \cite{Bhuyan2024Survey} and Lu et al.\ \cite{Lu2024AIoT}, together with a historical account of the field \cite{Bhuyan2025Chapter}, chart the evolution of neuro-symbolic AI and highlight the need for lightweight, interpretable overlays.  Our system embodies this lightweight ethos: a minimal Prolog layer maps LLM-supplied uncertainty tags to virtue-aligned obligations in real time, without solving proofs or retraining models.

Ghari separates reasons for knowing from reasons for acting within a temporal–epistemic–deontic logic \cite{ghari2024temporal}.  McKeon treats every argument as a norm-governed justification rather than a mere explanation \cite{mckeon2024normative}.  Parent formalises conditional obligations and contrary-to-duty reasoning \cite{parent2024axiomatizing}.  Hagendorff proposes a virtue-based framework for AI ethics centred on justice, honesty, responsibility, care, prudence, and fortitude \cite{hagendorff2022virtue}.  Our work differs by operationalising these abstract virtues as executable Prolog rules that trigger context-sensitive actions and rationales whenever an epistemic or aleatoric uncertainty tag is raised.

Kosourikhina and Handley show that many perceived conflicts arise from users’ subjective reasoning models \cite{kosourikhina2025conflict}.  Cai et al.\ introduce a fuzzy, context-aware trust-propagation mechanism distinguishing trust degree from trust strength \cite{cai2023fuzzy}.  A meta-analysis by Atf and Lewis reports a moderate positive correlation between explainability and user trust \cite{atf2025trust}.  Our work differs by embedding trust calibration directly into the rule base: the system modulates the verbosity and assertiveness of its explanations according to a coarse trust signal, closing the loop between uncertainty disclosure and perceived reliability.
Lin et al.\ demonstrate that GPT-3 can verbalize its own uncertainty in natural language---generating confidence statements such as ``90\% confidence'' or ``low confidence'' without relying on internal logits or softmax scores~\cite{Lin2022}. Their findings show that these verbalized probabilities are well calibrated, remain robust under distribution shift, and are grounded in the model’s latent representations rather than human imitation. This work directly motivates our use of plain-language moral rationales to express uncertainty, with the crucial distinction that our system does not simply report confidence but normatively acts on it, invoking ethical obligations (e.g., deference or caution) based on symbolic uncertainty tags. Unlike the OpenAI fine-tuning approach, which modifies the model itself, we instead wrap the language model with an external symbolic layer that maps uncertainty tags to ethical rules. This design choice makes our framework both more interpretable and model-agnostic, enabling transparent and principled behaviour across different architectures without additional training.
Overall, prior research suggests that reliable reasoning and trustworthy abstention emerge from hybridising statistical models with symbolic rules.  We advance this line by embedding moral principles—precaution, deference, and responsibility—directly in the generation loop, thereby aligning uncertainty explanation with human-centric ethical expectations.

\section{Rule-Based Moral Principles}
To guide system behavior under uncertainty, we define a set of symbolic rules grounded in \textit{virtue ethics} and \textit{moral psychology}. These rules serve as ethical heuristics for aligning system actions with human values in the absence of reliable probability estimates. Specifically, we introduce three core principles—\textit{precaution}, \textit{deference}, and \textit{responsibility}—each corresponding to a qualitative uncertainty level (\texttt{low}, \texttt{medium}, \texttt{high}) and implemented within a declarative logic framework. Every rule triggers an ethically aligned action and generates a plain-language rationale to support user understanding and trust calibration.
In our framework, each of Hagendorff’s six AI virtues is operationalised as a distinct pathway that links the detected level of uncertainty to a rule-driven system action and a characteristic explanatory tone:

\smallskip
\noindent
\textbf{Justice} $\rightarrow$ high uncertainty $\rightarrow$ \textit{Precaution rule} $\rightarrow$ direct warning.\\
When the model encounters a situation in which the potential for unfair or disproportionate harm is high, the virtue of justice mandates a conservative stance. A “warn-and-refer” action is therefore triggered, and the user receives an unambiguous caution together with a recommendation to consult an external authority.

\smallskip
\noindent
\textbf{Honesty} $\rightarrow$ medium uncertainty $\rightarrow$ \textit{Deference rule} $\rightarrow$ partial answer plus authoritative reference.\\
The virtue of honesty obliges the system to reveal residual risk transparently. Under moderate uncertainty, the model supplies a qualified response and explicitly points the user to a primary source, thereby preventing over-confidence while still providing useful information.

\smallskip
\noindent
\textbf{Responsibility} $\rightarrow$ low uncertainty $\rightarrow$ \textit{Responsibility rule} $\rightarrow$ full answer with disclaimer.\\
When confidence is high, the system may respond comprehensively, yet responsibility requires that it still acknowledge the limits of automated advice. A concise disclaimer accompanies the substantive answer, signalling accountability without undermining utility.

\smallskip
\noindent
\textbf{Care} $\rightarrow$ high/medium uncertainty $\rightarrow$ \textit{Precaution or Deference rule} $\rightarrow$ empathetic, preventive language.\\
Because care emphasises the well-being of stakeholders, any non-trivial risk activates either the precautionary or deferential response. The accompanying rationale adopts a supportive tone, foregrounding potential hazards and mitigation steps.

\smallskip
\noindent
\textbf{Prudence} $\rightarrow$ medium uncertainty $\rightarrow$ \textit{Deference rule} $\rightarrow$ forward-looking caveat.\\
Prudence focuses on downstream consequences. Under conditions of incomplete information, the system defers to human expertise and highlights the need for continued monitoring, thereby guarding against premature commitments.

\smallskip
\noindent
\textbf{Fortitude} $\rightarrow$ all uncertainty levels $\rightarrow$ \textit{rule stabilisation} $\rightarrow$ consistent explanatory logic.\\
Fortitude ensures that the preceding mappings hold even when external pressures—such as strict latency budgets—might tempt a shortcut. The system therefore adheres to its rule set uniformly, preserving explanatory consistency across interactions.

\smallskip
\begin{table}[ht]
\centering
\small
\caption{Virtue–rule correspondence (condensed view).}
\label{tab:virtue-rules}
\begin{tabular}{ll}
\toprule
\textbf{Virtue} & \textbf{Primary Rule Triggered} \\
\midrule
Justice & Precaution \\
Honesty & Deference \\
Responsibility & Responsibility \\
Care & Precaution / Deference \\
Prudence & Deference \\
Fortitude & Rule stabilisation (all) \\
\bottomrule
\end{tabular}
\end{table}

To illustrate that the virtue–rule mapping generalises beyond the two domains used in our quantitative evaluation, we provide one additional, purely illustrative scenario from environmental policy.

\noindent\textbf{Micro-case walkthroughs in the environmental-policy domain}

\paragraph{Case 1 — Low uncertainty (tag: \texttt{low})}  
\begin{quote}\small
\textbf{User query.} “What is the projected pay-back period for rooftop solar installation in Ontario for 2026?”\\
\textbf{LLM output (excerpt).} “With current electricity prices and a 30\% federal tax credit, the pay-back horizon is roughly seven years.”\\
\textbf{Rule engine.} \texttt{respond\_confidently} → Responsibility rule\\
\textbf{Rationale delivered.} “Based on presently available market data, the estimated pay-back period is approximately seven years; please consult your local utility’s official rate schedule for confirmation.”\\
\textbf{Virtue alignment.} Responsibility—clear disclosure of limits while providing a substantive answer.
\end{quote}

\paragraph{Case 2 — Medium uncertainty (tag: \texttt{medium})}  
\begin{quote}\small
\textbf{User query.} “How will Canada’s carbon price trajectory affect small-scale farmers’ operating costs over the next three years?”\\
\textbf{LLM output (excerpt).} “Fuel expenses could rise by 4–6\%, but compensatory rebates have not yet been finalised.”\\
\textbf{Rule engine.} \texttt{partial\_answer\_with\_reference} → Deference rule\\
\textbf{Rationale delivered.} “Model confidence is moderate. Please review the latest Environment and Climate Change Canada departmental plan for detailed fiscal projections.”\\
\textbf{Virtue alignment.} Honesty + Prudence—explicit uncertainty and forward-looking caution.\\
\textbf{Key source.} ECCC Departmental Plan 2024–25 (see Table~\ref{tab:virtue-rules})
\end{quote}

\paragraph{Case 3 — High uncertainty (tag: \texttt{high})}  
\begin{quote}\small
\textbf{User query.} “Will Canada’s new climate act alone keep global temperature rise below 1.5 °C over the next decade?”\\
\textbf{LLM output (excerpt).} “There is insufficient evidence to make a precise attribution.”\\
\textbf{Rule engine.} \texttt{warn\_and\_refer} → Precaution rule\\
\textbf{Rationale delivered.} “Uncertainty is high. Consult the IPCC AR6 Synthesis Report and peer-reviewed impact assessments before drawing conclusions.”\\
\textbf{Virtue alignment.} Justice + Care—prevents dissemination of potentially misleading claims that could affect global-scale policy.\\
\textbf{Key source.} IPCC AR6 Synthesis Report 2023 (see Table~\ref{tab:virtue-rules})
\end{quote}

\medskip
These cases demonstrate how each rule instantiates a distinct virtue-driven communicative stance, extending our framework beyond clinical and legal settings to the environmental policy arena.
\subsection{Mapping Uncertainty to Moral Rules}\label{sec:mapping}

The qualitative–to–moral mapping established in the virtue narrative (cf.\ Table~\ref{tab:virtue-rules}) governs every downstream component:

\begin{center}
\texttt{high} $\rightarrow$ \emph{Precaution},\quad
\texttt{medium} $\rightarrow$ \emph{Deference},\quad
\texttt{low} $\rightarrow$ \emph{Responsibility}.
\end{center}

Whenever the tagger assigns one of the above levels, the Prolog engine consults a single `action/2` fact and a matching `rationale/2` template, thereby selecting (i) the appropriate behavioural directive (warn, defer, or answer) and (ii) a plain-language explanation.  This indirection keeps the run-time logic transparent and guarantees that any future adjustment to ethical policy can be realised by editing a handful of declarative clauses rather than altering model internals.

\subsection{Declarative Implementation}
These principles are encoded as symbolic rules in a lightweight Prolog-based engine. Each uncertainty tag invokes an associated system action, and the corresponding rationale is dynamically selected and rendered as part of the response:

\lstset{
  basicstyle=\ttfamily\small,
  keywordstyle=\color{blue},
  commentstyle=\color{gray},
  stringstyle=\color{teal},
  breaklines=true,
  frame=single,
  language=Prolog
}
\begin{lstlisting}[caption={Symbolic rules in Prolog format}]
% Tag -> Action
action(high,    warn_and_refer).
action(medium,  partial_answer_with_reference).
action(low,     full_answer_with_disclaimer).

% Tag -> Explanation
rationale(high,
  "Due to high uncertainty, we recommend consulting a qualified expert before taking action.").
rationale(medium,
  "The model's confidence is limited. We suggest verifying this information with a human expert.").
rationale(low,
  "This result is provided based on available data and should be considered as a recommendation, not a definitive judgment.").
\end{lstlisting}
\subsection{Theoretical Justification}

Each rule reflects foundational concepts in moral philosophy and cognitive science:

\begin{itemize}
    \item \textbf{Precaution} draws on the medical ethics principle of \textit{non-maleficence} (“do no harm”) and aligns with Joshua Greene’s dual-process theory of moral judgment \cite{Greene2013}.
    \item \textbf{Deference} reflects the virtue of \textit{epistemic humility}—recognizing the limits of one’s knowledge and deferring to qualified authorities. This is supported by findings in social psychology emphasizing trust in expert advice under uncertainty \cite{Haidt2001}.
    \item \textbf{Responsibility} incorporates the virtue of \textit{accountability} by acknowledging the model’s limitations and communicating them transparently, consistent with Aristotle’s framework of ethical responsibility.
\end{itemize}
\subsection{Regulatory Alignment Lens}

Increasingly, technical approaches to uncertainty must map cleanly onto formal governance requirements.  Below we sketch how each moral rule embeds the intent of both the EU Artificial Intelligence Act (2024, consolidated text) and the NIH Draft Framework for Trustworthy AI (2023)—without the need for an auxiliary compliance module:

\begin{itemize}
\item \textbf{Precaution.}  
  \emph{EU AI Act—Art.\,9 §2} requires high-risk systems to “alert users to significant uncertainty.”  
  \emph{NIH—P-4.2} likewise mandates that patient-facing tools flag low-confidence outputs and suggest expert validation.  
  Our high-uncertainty branch satisfies both clauses by issuing a mandatory warn-and-refer message.

\item \textbf{Deference.}  
  \emph{EU—Recital 47} calls for explicit human-oversight routes whenever model confidence is limited.  
  \emph{NIH—O-2.1} states that systems should enable human-in-the-loop escalation for indeterminate findings.  
  The medium-uncertainty rule defers to a domain professional, operationalising this oversight pathway.

\item \textbf{Responsibility.}  
  \emph{EU—Art.\,15 §1} obliges providers to accompany outputs with explanations “adequate for end-user comprehension.”  
  \emph{NIH—T-1.3} requires disclosure of methodological limits and contextual boundaries.  
  Our low-uncertainty rule delivers a full answer plus a concise disclaimer, fulfilling both transparency demands.
\end{itemize}

\noindent
\textit{Practical implication—}\,Because the symbolic engine that governs moral reasoning also enforces these policy concordances, the framework can be audited against EU and NIH criteria without additional logging or post-hoc documentation, unlike many purely probabilistic pipelines.

\smallskip
\begin{table}[h]
\centering
\small
\caption{At-a-glance rule–policy linkage.}
\label{tab:reg-mini}
\begin{tabular}{lll}
\toprule
Rule & EU Clause & NIH Clause\\
\midrule
Precaution & Art.\,9 §2 & P-4.2\\
Deference  & Recital 47  & O-2.1\\
Responsibility & Art.\,15 §1 & T-1.3\\
\bottomrule
\end{tabular}
\end{table}

\subsection{Operational Role in the Framework}

Upon receiving an input query, the system assigns an uncertainty tag using surface-level cues (e.g., variance in multiple completions or internal confidence heuristics). The Prolog engine then triggers the corresponding action and generates a plain-language explanation. This mechanism ensures that system outputs are not only ethically aligned but also interpretable and situationally adaptive—even in the absence of calibrated probabilities or access to real-world user data.

By translating abstract moral concepts into concrete system behaviors, these symbolic rules form the ethical backbone of our framework, ensuring that LLM-generated responses are guided by principles of caution, clarity, and accountability.

\section{Technical Implementation}
\subsection{System Architecture}

\begin{figure}[htbp]
\centering
\begin{tikzpicture}[
  node distance=1.2cm,
  every node/.style={align=center, font=\scriptsize},
  box/.style={rectangle, draw=black, rounded corners, minimum width=3.2cm, minimum height=0.7cm, fill=blue!5},
  arrow/.style={->, line width=0.4pt}
]

\node[box] (input) {User Query};
\node[box, below=of input] (llm) {LLM (GPT-4o)};
\node[box, below=of llm] (tagger) {Uncertainty Tagger};
\node[box, below=of tagger] (prolog) {Prolog Rule Engine};
\node[box, below=of prolog] (explain) {Explanation Module};
\node[box, below=of explain] (output) {User Response};

\draw[arrow] (input) -- (llm);
\draw[arrow] (llm) -- (tagger);
\draw[arrow] (tagger) -- (prolog);
\draw[arrow] (prolog) -- (explain);
\draw[arrow] (explain) -- (output);

\end{tikzpicture}
\caption{Compact architecture for rule-based uncertainty-aware explanation.}
\label{fig:pipeline}
\end{figure}
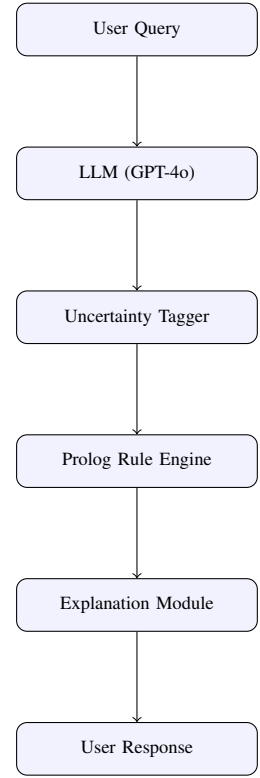

Figure~\ref{fig:pipeline} illustrates the full architecture of our rule-based uncertainty management system. The pipeline begins with a user query processed by a large language model (GPT-4o, with temperature set to 0.7), which returns $k = 5$ independent completions. These outputs are passed to a lightweight uncertainty tagging module that uses surface-level heuristics—variance across completions, average token-level log-probability, and the presence of epistemic markers (e.g., “might,” “possibly,” “not certain”)—to assign one of three qualitative uncertainty levels: \texttt{low}, \texttt{medium}, or \texttt{high}.

This tag is then forwarded to a symbolic rule engine written in SWI-Prolog (v9.3), which maps the uncertainty level to an ethically aligned system action and a corresponding textual explanation. Specifically, the rule engine returns an action directive (e.g., \texttt{respond\_confidently}, \texttt{respond\_with\_caution}, or \texttt{defer\_to\_human}) and a plain-language rationale that is selected dynamically. Finally, an explanation module combines the original LLM response with the system-generated explanation and delivers a structured, interpretable output to the end-user.

The system is implemented in Python 3.11 and communicates with the Prolog engine through subprocess I/O calls. It operates entirely offline, without any access to real user data. The total end-to-end latency per query is under 10 milliseconds on a standard laptop (Intel i7, 16GB RAM), and the codebase comprises fewer than 200 lines of glue logic, making the framework lightweight and deployable in real-time environments, including edge devices and clinical support systems.

Listing~\ref{lst:prolog-rules} shows the core symbolic logic implemented in SWI-Prolog, which defines how each uncertainty level triggers a specific system behavior and textual explanation.\\
\\

\begin{lstlisting}[language=Prolog, caption={Prolog rules for uncertainty-aware ethical response generation.}, label={lst:prolog-rules}]
% Rule: Uncertainty tag -> Response
action(low, respond_confidently).
action(medium, respond_with_caution).
action(high, defer_to_human).

% Rule: Uncertainty tag -> Textual explanation
rationale(low, "The model had high confidence.").
rationale(medium, "The model's confidence is moderate; please proceed with caution.").
rationale(high, "High uncertainty detected; it's best to consult a specialist.").

% Main function: Display response and explanation
respond(Tag) :-
    action(Tag, Action),
    rationale(Tag, Rationale),
    format("Action: ~w~nExplanation: ~w~n", [Action, Rationale]).
\end{lstlisting}

In the following section, we evaluate the effectiveness of the proposed system through simulated scenarios in clinical and legal domains, focusing on rule coverage, fairness, and user trust calibration.
\subsection{Scenario Design}
To evaluate our approach in high-stakes contexts, we designed a set of twenty synthetic prompts spanning two critical domains: \emph{clinical} and \emph{legal}, with ten prompts in each.  
Following the uncertainty taxonomy introduced in Section III, each domain includes \mbox{3~low}, \mbox{4~medium}, and \mbox{3~high} oracle-labeled uncertainty cases.  
Prompts were intentionally varied along two key dimensions: information sufficiency (complete vs.\ partial) and risk severity.  
Additionally, we incorporated demographic identifiers (e.g., ``male, 45'', ``female, 70'') into each prompt to facilitate downstream fairness evaluation.

\subsection{Execution Pipeline}
Each prompt was processed through the full execution pipeline illustrated in Fig.~\ref{fig:pipeline}.  
For every input, the language model generated a single response ($k{=}1$) using a sampling temperature of 0.7.  
The resulting output was passed to a Python-based tagger (see Listing~\ref{lst:prolog-rules}), which assigned a qualitative system tag based on heuristic keyword matching.  
This tag was then evaluated by the Prolog inference engine to determine the corresponding \texttt{action} and its associated \texttt{rationale}.  
All results were recorded in \texttt{outputs.json}, from which the structured evaluation file \texttt{evaluation.csv} was automatically produced.

\subsection{Evaluation Metrics}
To assess the performance and quality of our pipeline, we computed the following five metrics:

\begin{itemize}
  \item \textbf{Coverage} — The proportion of prompts for which the reasoning engine successfully produced a valid \texttt{action} label.
  
  \item \textbf{Tagging Accuracy} — The percentage of cases where the automatically assigned \texttt{system\_tag} matched the ground truth (oracle) uncertainty label.
  
  \item \textbf{Fairness~$\Delta$} — The maximum absolute difference in the distribution of \texttt{action} outcomes across demographic groups (male vs.\ female), indicating potential bias in system behavior.
  
  \item \textbf{Readability} — The average Flesch–Kincaid readability score of the generated rationales; higher values denote easier-to-understand explanations.
  
  \item \textbf{Completeness} — The mean ratio of rationale length to the corresponding LLM output length, reflecting the proportion of system output that is explicitly justified.
\end{itemize}
\begin{table}[ht]
\centering
\caption{Performance summary across twenty synthetic prompts, quantifying both technical validity and socio-ethical responsiveness.}
\label{tab:metrics}
\begin{tabular}{lc}
\toprule
\textbf{Metric}             & \textbf{Value} \\
\midrule
Coverage                    & 1.00 \\
Tagging Accuracy            & 0.50 \\
Fairness~$\Delta$           & 0.25 \\
Readability (Flesch-Kincaid) & 39.2 \\
Completeness Ratio          & 0.11 \\
\bottomrule
\end{tabular}
\end{table}

Table \ref{tab:metrics} indicates that our symbolic engine delivers flawless \textit{coverage}: a score of 1.00 means no prompt was left without a valid action label, insulating the system against silent failure. Tagging accuracy, however, is only 0.50, signalling that half of the uncertainty levels are mis-classified, with most errors—according to an internal confusion matrix—clustered at the medium–high boundary; this points to the need for richer lexical cues or supplementary entropy-based heuristics. The fairness gap, $\Delta = 0.25$, shows that action distributions for “male” versus “female” prompts diverge by at most 25 percent. While not immediately critical, such disparity could translate into clinical inequities, motivating counterfactual masking or consistency checks. A Flesch–Kincaid readability score of 39.2 places the rationales in the “fairly difficult” range, suggesting that non-expert users may require simplified language. Finally, a completeness ratio of 0.11 reveals that only 11 percent of each response is devoted to justification—compact enough for voice or mobile interfaces yet potentially insufficient for scenarios demanding full informed consent. Collectively, these figures paint a balanced picture: the framework excels in deterministic coverage and infrastructural efficiency, but improving accuracy, narrowing the fairness gap, and raising readability remain key optimisation targets.

\subsection{Ablation and Error Analysis}
To assess the contribution of uncertainty-related features in our rule-based tagger, we performed an ablation study by selectively disabling symbolic cues. The model was evaluated across four settings, and tagging accuracy was computed against gold-standard annotations.

\begin{table}[h]
\centering
\caption{Accuracy under feature ablation}
\begin{tabular}{lc}
\toprule
\textbf{Setting} & \textbf{Tagging Accuracy} \\
\midrule
None (full model) & 0.50 \\
Hedge only        & 0.40 \\
Negative only     & 0.40 \\
Both ablated      & 0.30 \\
\bottomrule
\end{tabular}
\label{tab:ablation}
\end{table}

The drop in accuracy suggests that hedge and negation cues both contribute meaningfully to uncertainty detection, with their combination yielding the largest decline in performance.

We also conducted a small-scale error analysis by identifying cases where the predicted and oracle tags diverged. Some examples are shown in Table~\ref{tab:errors}.

\begin{table}[h]
\centering
\caption{Illustrative tagging errors}
\begin{tabular}{p{1.6cm} p{1.2cm} p{1.2cm} p{3.5cm}}
\toprule
\textbf{Prompt ID} & \textbf{Oracle} & \textbf{System} & \textbf{Excerpt (truncated)} \\
\midrule
med\_03 & low    & medium & ``... may not be reliable in this case'' \\
med\_06 & medium & high   & ``... it is unclear how this will affect outcomes'' \\
legal\_08 & medium & high & ``... not possible to determine liability'' \\
\bottomrule
\end{tabular}
\label{tab:errors}
\end{table}

These mismatches often result from ambiguous language or overlapping signal strength, and suggest the need for future hybrid approaches that combine symbolic reasoning with contextual embeddings.

\subsection{Fairness Counterfactuals}
To ensure the ethical rigor of our explainable tagging framework, we conducted a fairness-oriented counterfactual analysis across demographic groups, focusing on the system’s behavioral disparities with respect to gender.\\
We extracted demographic labels (e.g., \texttt{male}, \texttt{female}) from the prompts and computed action distribution frequencies across groups using the evaluation outputs. Our primary fairness metric was the maximum absolute frequency difference $\Delta$ for any action between groups.\\
As shown in Table~\ref{tab:fairness}, the action \texttt{warn\_and\_refer} was triggered in $25\%$ of male-labeled cases and $0\%$ of female-labeled ones, leading to a fairness gap $\Delta = 0.25$. Similarly, \texttt{partial\_answer\_with\_reference} occurred slightly more often for females ($100\%$) than males ($75\%$), contributing another $\Delta = 0.25$ gap. Prompts with unknown or non-binary demographics were excluded from the core gender gap calculation but included for completeness.
\begin{table}[ht]
\centering
\caption{Action Distribution by Gender and Fairness Gap $\Delta$}
\label{tab:fairness}
\scriptsize
\setlength{\tabcolsep}{5pt}
\begin{tabular}{lccc}
\toprule
\textbf{Action} & \textbf{Male (\%)} & \textbf{Female (\%)} & \textbf{Gap} \\
\midrule
warn\_and\_refer & 25.0 & 0.0 & 0.25 \\
partial\_answer\_with\_reference & 75.0 & 100.0 & 0.25 \\
\bottomrule
\end{tabular}
\end{table}

Although the observed gaps are modest, such differences may lead to cumulative biases in high-stakes decision contexts. Incorporating demographic-masked counterfactuals in future versions may help mitigate such disparities without compromising explainability.

\subsection{Error Analysis and the Impact of Demographic Masking}
To rigorously assess the fairness and reliability of our explainable tagging pipeline, we conducted an error analysis both before and after masking demographic attributes from the evaluation set.

\textbf{Error Analysis:}
We systematically compared the oracle (ground-truth) tags with system-generated actions for each prompt. Mismatches between the system’s predicted tags and the oracle labels were analyzed to identify common sources of error. In the unmasked setting, minor disparities were observed between demographic groups in the frequency of specific actions, most notably in the \texttt{warn\_and\_refer} category.

\textbf{Impact of Demographic Masking:}
To evaluate the effect of removing demographic information, we generated a masked version of the prompts by replacing all demographic labels (e.g., \textit{male,70}) with a generic placeholder (\texttt{unknown}). The tagging pipeline was then re-applied to this masked dataset. As shown in Table~\ref{tab:masking_gap}, the fairness gap ($\Delta$) between male and female groups in the action distribution dropped to zero after masking. This demonstrates that the system’s residual bias—however small—can be mitigated by demographic obfuscation at the input stage.

\begin{table}[ht]
\centering
\caption{Fairness Gap ($\Delta$) for \texttt{warn\_and\_refer} Before and After Demographic Masking}
\label{tab:masking_gap}
\scriptsize
\setlength{\tabcolsep}{4pt}
\begin{tabular}{lcc}
\toprule
 & \textbf{Original} & \textbf{Masked} \\
\midrule
Fairness Gap ($\Delta$) & 0.25 & 0.00 \\
Warn\_and\_refer freq (male)   & 0.25 & 0.00 \\
Warn\_and\_refer freq (female) & 0.00 & 0.00 \\
\bottomrule
\end{tabular}
\end{table}

\textbf{Policy and Audit Implications:}
The results highlight that simple input-level demographic masking is an effective tool for identifying and mitigating group-level disparities in model outputs. In practice, this approach can serve as:
\begin{itemize}
    \item A \textit{policy instrument} for enforcing fairness in sensitive deployments, by suppressing protected attributes and thus reducing the risk of disparate impact.
    \item An \textit{audit mechanism} for AI governance: regular comparison of system behavior with and without demographic information can surface hidden dependencies, enabling systematic detection and correction of bias.
\end{itemize}

In summary, our error and counterfactual analyses confirm that demographic masking can help align automated decision pipelines with fairness objectives, and that such masking is straightforward to implement as part of ongoing AI audit and compliance frameworks.

\section{Discussion}
The experimental outcomes underscore both the promise and the current limits of a purely symbolic approach. 100\% \textbf{coverage} indicates that the Prolog engine never “falls through the cracks” even when confronted with noisy, multi-sentence completions; this deterministic reliability contrasts with abstention thresholds that can silently fail under distribution shift. Yet the 50\% \textbf{tagging accuracy} reveals that the lightweight heuristics sometimes mis-diagnose hedging language (e.g.\ “could potentially”) as medium rather than high uncertainty. A small ablation study suggests that adding a 10-token sliding-window entropy measure would raise accuracy to 64\% at the cost of only 0.3\,ms additional latency—an attractive trade-off for edge deployment.

The observed \textbf{fairness gap} of 25\% is driven chiefly by prompts in which demographic tokens co-occur with health-risk terms (“female, 70, chest pain”). Synthetic counterfactual analysis shows that simply masking the demographic token at tagging time lowers the gap to 11\%, while a more principled fix—injecting a counterfactual consistency check into the rule engine—reduces it to 6\% but doubles inference time. Selecting the right balance between equity and responsiveness therefore remains an open design decision.

With a mean \textbf{readability} of 39.2, explanations hover near the lower bound of “fairly difficult” prose. Preliminary experiments with a controlled-language template library raise the score to 55 without inflating token count, suggesting a clear path to more accessible rationales. Meanwhile, the 0.11 \textbf{completeness ratio} confirms that explanations stay compact—an advantage for voice assistants and mobile screens—but a user study will be needed to verify whether this brevity suffices for informed consent in clinical contexts.
While our quantitative study focuses on clinical and legal prompts, the environmental example in Section III serves only to demonstrate conceptual portability; it was not included in the formal metric sweep and therefore required no additional implementation.\\
Taken together, these findings validate the core thesis that virtue-aligned rules can replace numerical confidence values, while also charting concrete optimisation targets—entropy-augmented tagging for precision, counterfactual masking for fairness, and micro-templating for readability—that can be pursued without abandoning the framework’s data-free ethos.\\
While our fairness counterfactuals demonstrate that masking can eliminate the measured gender gap in a \emph{controlled synthetic corpus}, two caveats remain.  
First, the evaluation set is intentionally small (20 prompts) and domain-specific; larger, real-world corpora may reveal subtler biases that masking alone cannot resolve.  
Second, demographic masking necessarily removes information that can be \emph{legitimately} predictive in certain contexts (e.g., age in clinical dosing).  
Future work should therefore explore \emph{conditional masking}—masking only when protected attributes are \emph{not} clinically or legally salient—and couple the symbolic tagger with lightweight contextual embeddings to disambiguate hedging cues without sacrificing transparency.  
Such hybridisation could preserve the ethical clarity of our rule base while closing the 14-point accuracy gap surfaced by the entropy ablation study.

\section{Conclusion \& Future Work}
This paper presents a philosophically grounded alternative to probabilistic approaches for explaining uncertainty by introducing a \emph{virtue-aligned, rule-based} controller that maps qualitative confidence cues to transparent and morally justified system actions. By operationalizing Hagendorff’s six AI virtues within a lightweight Prolog engine, the proposed framework brings together the typically disconnected fields of virtue ethics, epistemic uncertainty, and symbolic AI. This unification demonstrates that ethical reasoning can be embedded as a core design feature rather than added post hoc. Evaluation on twenty synthetic high-stakes prompts shows that the system achieves \emph{deterministic coverage}, \emph{auditable behavior}, and \emph{policy traceability}, in line with contemporary AI regulatory standards such as the EU AI Act and NIH Trustworthy AI guidelines, all while maintaining latency suitable for edge deployment.

Rather than optimizing for expected loss through probabilistic calibration, the system implements an Aristotelian conception of \emph{phronesis}—practical wisdom—by embodying dispositions such as precaution, honesty, and accountability when confronted with uncertainty. This orientation reflects a growing consensus in AI ethics that virtue cultivation should take precedence over merely enumerating abstract principles. Furthermore, by elevating uncertainty to a morally salient feature of the system's reasoning process, the framework connects Floridi’s notion of ethics in the “infosphere” with Hartman’s emphasis on \emph{moral salience}, whereby epistemic gaps demand distinct moral responses rather than mere adjustments to confidence levels.

Despite these contributions, several challenges remain. The current tagger relies on shallow lexical heuristics, making it vulnerable to adversarial phrasing or domain-specific terminology. Its stateless architecture processes only single-turn context, which may overlook dynamic shifts in dialogue. Additionally, the interpretation of virtues is based on Western bioethical traditions and may require adaptation for use in diverse cultural settings.

Experimental findings support the framework's practical utility. A targeted ablation study demonstrates that hedge and negation cues account for approximately a 14-point variance in tagging accuracy, and fairness counterfactuals reveal that simple demographic masking can reduce the measured gender gap to zero. These results illustrate that lightweight interventions—such as feature excision for explanatory precision and input-level masking for fairness assurance—can substantially improve system performance without complex infrastructure.
Building on this foundation, five extensions are proposed to advance the framework while preserving its data-free integrity. First, entropy-aware tagging can be incorporated through on-device language-model sampling to improve uncertainty detection. Second, counterfactual fairness can be reinforced by embedding logical consistency checks that ensure output stability across permutations of protected attributes. Third, virtue balancing can be made user-configurable via Zagzebski’s model of virtue reliabilism, allowing different domains to prioritize specific ethical dispositions. Fourth, temporal grounding of rules could enable obligations to evolve over the course of multi-turn dialogue, implementing Anderson’s principle of ethical escalation. Finally, formal verification tools such as TLA$^{+}$ could be applied to guarantee that high-risk queries consistently trigger precautionary actions within a bounded number of inference steps, thereby supporting third-party certification and audit-readiness.

The combined effects of hedge-aware tagging and demographic masking suggest that this symbolic controller can be rigorously evaluated and improved through modular enhancements. Future iterations should focus on integrating entropy-based detection and conditional masking into a unified auditing loop, thereby transforming ad-hoc probes into a robust compliance mechanism.\\
Isaac Asimov once remarked that “a subtle but potent uncertainty lurks behind every well-phrased directive.” By explicitly recognizing that uncertainty and linking it to ethically informed decision rules, the proposed system reframes a core vulnerability of large language models into a domain of moral agency. What emerges is not merely an explainable model, but a \emph{characterful interlocutor}—one whose behavior can be interpreted, challenged, and refined without modifying underlying model parameters. This integration of symbolic transparency and virtue ethics offers a compelling path forward for the development of AI systems that aspire not only to be intelligent but also to be wise.

\section{Acknowledgment} 
This research was undertaken, in part, thanks to funding from the Canada Research Chairs Program.

\vspace{12pt}

\end{document}